\documentclass{article}

\PassOptionsToPackage{numbers, sort}{natbib}

\usepackage[main, final]{neurips_2026}

\makeatletter
\DeclareRobustCommand\onedot{\futurelet\@let@token\@onedot}
\def\@onedot{\ifx\@let@token.\else.\null\fi\xspace}

\def\eg{\emph{e.g}\onedot,~} 
\def\ie{\emph{i.e}\onedot,~}

\makeatother

\usepackage[utf8]{inputenc}
\usepackage[T1]{fontenc}
\usepackage{url}
\usepackage{booktabs}
\usepackage{amsmath}
\usepackage{amssymb}
\usepackage{amsfonts}
\usepackage{nicefrac}
\usepackage{microtype}
\usepackage[table]{xcolor}
\definecolor{refblue}{rgb}{0.21,0.49,0.74}
\usepackage[pagebackref,breaklinks,colorlinks,allcolors=refblue]{hyperref}

\usepackage{graphicx}
\graphicspath{{images/}}
\usepackage{xspace}
\usepackage{enumitem}
\usepackage{caption}
\usepackage{multirow}
\usepackage{duckuments}
\usepackage{nicematrix}
\usepackage{pifont}
\newcommand{\cmark}{\ding{51}}%
\newcommand{\xmark}{\ding{55}}%

\usepackage{kotex}

\usepackage[capitalize]{cleveref}
\crefname{section}{Section}{Sections}
\crefname{table}{Table}{Tables}
\crefname{figure}{Figure}{Figures}
\crefname{equation}{Equation}{Equations}
\crefname{appendix}{Appendix}{Appendices}

\newcommand{\method}{{Pose6DAug}\xspace}

\newcommand{\zero}{0.0}
\newcommand{\relative}{16.5}

\title{Pose6DAug: Physically Plausible Multi-view Object Swapping for Robot Data Augmentation}

\author{%
Jonghoon Lee\textsuperscript{1}\thanks{Equal contribution.}\enspace
Seong Hyeon Park\textsuperscript{2}\footnotemark[1]\enspace
Byungwoo Jeon\textsuperscript{1}\enspace
Minha Lee\textsuperscript{1}\enspace
Jinwoo Shin\textsuperscript{1,3}\thanks{Corresponding author.}\\[4pt]
\normalfont\textsuperscript{1}KAIST\quad \textsuperscript{2}Ajou University\quad \textsuperscript{3}RLWRLD\\[3pt]
\normalfont\texttt{\{alphabet1, jinwoos\}@kaist.ac.kr}, \texttt{seonghyp@ajou.ac.kr}%
}

\begin{document}

\maketitle

\begin{abstract}
Vision-language-action (VLA) policies have shown strong potential for general-purpose manipulation, yet they often fail on novel, out-of-distribution objects whose appearance or geometry deviates from the training distribution. 
The standard remedy is to collect multi-view teleoperation data for every failure case, but this scales poorly in both cost and time.
We introduce \method, a failure-driven data augmentation framework that turns existing successful episodes into targeted demonstrations for a policy's failure modes, without any new data collection.
Our key insight is that each successful episode already encodes a physically valid action trajectory together with calibrated multi-view observations. 
By swapping only the manipulated object while preserving this trajectory, we obtain new and physically grounded demonstrations. 
However, the challenge lies in synthesizing object swaps, \eg naive 2D video editing breaks multi-view consistency and physical plausibility, particularly under heavy occlusion and egocentric viewpoints.
Our method instead operates directly in 3D, anchoring the target object with an explicit mesh driven by a temporally coherent 6D pose trajectory.
This ensures geometrically consistent renderings across all camera views.
Fine-tuning a VLA on data augmented by our method improves success rates by \relative\% relative to the state-of-the-art baseline on novel objects, while preserving in-distribution performance. On a real robot, it more than doubles the success rate of the base policy.
These results show that multi-view and physically consistent augmentation is a practical path to scalable VLA generalization.

\noindent Project page: \url{https://jhoonjwa.github.io/pose6daug}
\end{abstract}

\section{Introduction}
\label{sec:intro}

\begin{figure}[t]
    \centering
    \includegraphics[width=\linewidth]{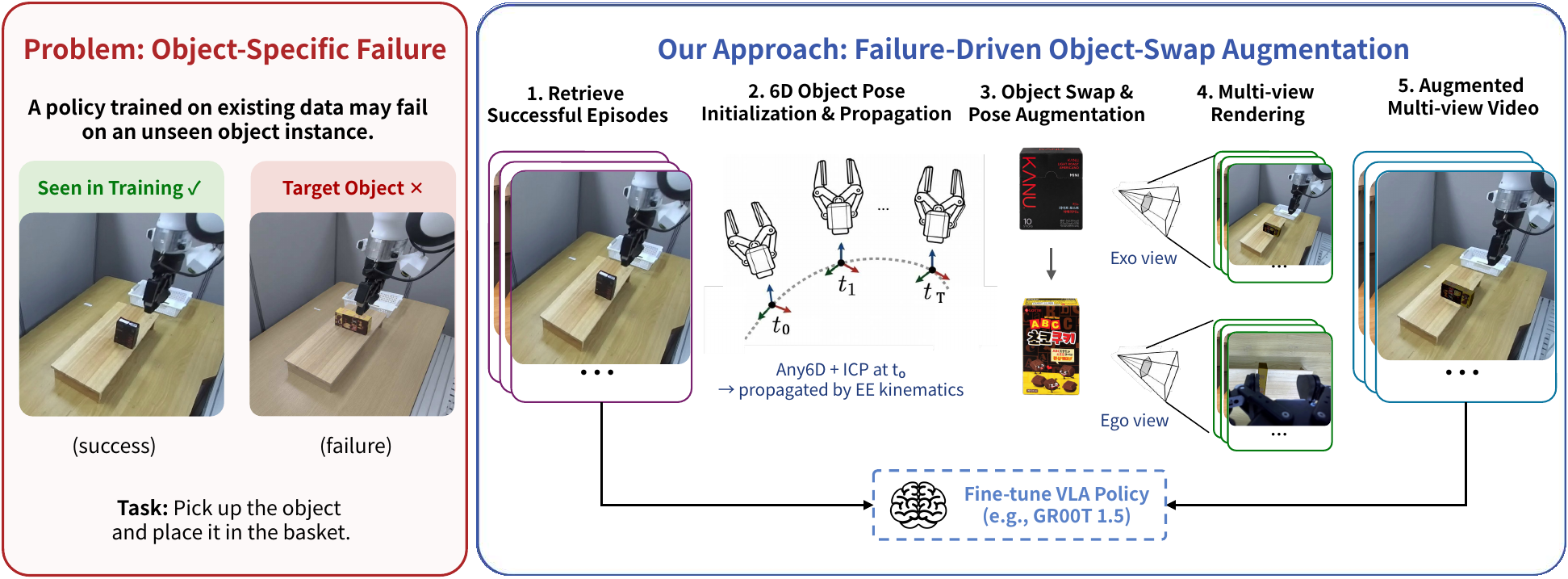}
    \caption{
    \textbf{Failure-driven object-swap augmentation.}
    {(a)} Object-specific failure where a base policy fails to roll out plausible actions on an episode with novel object instances.
    {(b)} Given such failed episodes, we aim to synthesize physically plausible demonstrations for their object instances (target objects).    
    We first retrieve successful demonstrations for the episodes with in-distribution object instances. Then, preserving their original action trajectory, we replace the manipulated objects with our target objects.
    Specifically, our method estimates a 3D mesh and a 6D pose sequence for a target object based on robot kinematics under a rigid coupling, which enables rendering physically plausible, multi-view consistent videos for fine-tuning the base policy.
    }
    \label{fig:main_pipeline}
\end{figure}
Vision-Language-Action (VLA) policies \citep{zitkovich2023rt, kim2024openvla, pi0, intelligence2025pi_0.5, groot, team2024octo, kim2026rldx} have shown significant promise in general-purpose manipulation, supported by a growing body of work that extends them along several axes, such as enriching their input modalities with tactile and force feedback~\citep{moss, forcevla} or strengthening the visual reasoning of their multimodal backbone~\citep{valr}. However, these models often suffer from a performance gap when deployed on novel object instances due to variations in appearance and geometry. While fine-tuning on new demonstrations can mitigate this issue, collecting high-quality multi-view robot data for every failure case is prohibitively expensive, creating a bottleneck for scaling robot learning~\citep{vlapilot, julg2025refined, liberopro}.

To alleviate the cost of data collection, prior works have explored various synthesis strategies, yet each presents critical trade-offs. Simulation-based trajectory adaptation~\citep{mimicgen, softmimicgen} can generate valid motions, but finding physically admissible samples for complex tasks often requires exhaustive search or rejection sampling, as well as manual segmentation of each demonstration into subtask segments.

Recently, with the advancement of video generative models~\citep{vace, wan, ltx, void}, it is possible to  synthesize visual counterfactuals for a single viewpoint. However, since they operate without explicit knowledge of the underlying 3D trajectory and apply edits in a view-independent manner, they are prone to producing physically implausible augmentations~\citep{robotransfer}. Moreover, this view-agnostic formulation precludes consistent augmentation across egocentric and exocentric viewpoints, a property essential for VLA models~\citep{groot, pi0} that learn from multiple views of the trajectory during training.

To address these limitations, we introduce \textbf{\method} (\cref{fig:main_pipeline}), a 3D-aware data augmentation framework that synthesizes targeted counterfactual demonstrations from existing successful episodes. Our core insight is that successful manipulation trajectories can be repurposed for novel objects through explicit 3D-anchored substitution. By operating in 3D space, \method bridges the gap between kinematic validity and visual photorealism, ensuring structural coherence across all observations.

Specifically, our method utilizes an explicit 3D representation to maintain spatial and temporal consistency across multiple camera views. Given a target object, we first extract its 3D mesh using an external perception model (\eg SAM3D \citep{sam3d}). We then erase the original object using its segmentation mask and restore the background via inpainting \citep{void}.
Subsequently, to ensure the synthesized object motion is strictly consistent with the robot's kinematics, we represent the interaction as a temporally coherent 6D pose trajectory.
By rendering the target mesh into the inpainted frames based on this kinematically consistent pose, our approach maintains geometric alignment as well as plausible object trajectories in the 3D space. Furthermore, to enhance data diversity, we inject physically valid geometric perturbations, such as object rotations, translations along the approach axis, and rescaling, which remain grounded in the 3D space.

We evaluate \method both in simulation, on the RoboCasa365 \citep{robocasa365} benchmark, and on a real robot. In RoboCasa365, it provides consistent performance gains across diverse novel object instances, achieving a relative \relative\% improvement in success rate and surpassing baselines that rely on 2D augmentation \citep{vace} or limited demonstration sets \citep{mimicgen}. On the real robot, \method more than doubles the base policy's success rate on novel objects, while 2D video editing fails to yield measurable performance gains.
Our results highlight the importance of 3D consistency and physical admissibility in robust policy learning.

\section{Related Work}
\label{sec:related_work}
\subsection{Data Augmentation for Robot Learning}
\label{subsec:data_aug}
Collecting large-scale demonstration data remains a bottleneck in robot learning due to the cost and time required for human teleoperation. Image-based augmentations using diffusion models~\citep{rosie, genaug} have gained increasing attention as a scalable alternative. These methods leverage text-to-image diffusion models to augment existing robot datasets by inpainting objects into recorded video frames. However, they apply edits independently per frame and per view, without explicit knowledge of the underlying 3D trajectory, and thus do not ensure geometric consistency across multi-view observations.

On the other hand, simulation-based approaches such as MimicGen~\citep{mimicgen} and its variants~\citep{softmimicgen} operate directly in a simulator to synthesize new demonstrations by adapting source end-effector trajectories via $SE(3)$ transformations to novel object poses. While capable of generating physically plausible 3D trajectories, these methods require manual annotation of subtask segments within each demonstration and depend on successful rollout execution in simulation. For difficult object geometries, valid rollouts may never be achieved within a feasible number of attempts, limiting both coverage and scalability.

\subsection{Visual Generative Models and Video Editing}
\label{subsec:video_editing}
The advancement of diffusion models has enabled various video editing and synthesis applications. Recent unified frameworks, such as {VACE}~\citep{vace}, integrate multiple video creation and editing tasks, including masked video-to-video editing, inpainting, and object swapping, into a single model conditioned on multimodal inputs. While these 2D generative models produce plausible edits in single-view scenarios, they lack explicit 3D awareness. Consequently, when applied to the rigid geometric constraints of robotic manipulation, they struggle to maintain precise physical admissibility, such as exact contact points during a grasp, and fail to preserve structural coherence across frames under the complex occlusions caused by robot grippers.

\subsection{Multi-View Consistency in Robotic Vision}
\label{subsec:multi_view}
For VLA policies to generalize effectively, they require multi-view training data that maintains strict spatial and temporal coherence. Applying 2D video editing models view-by-view introduces cross-camera inconsistencies in object scale, position, and appearance, which degrade policy learning. \method integrates the principles of kinematic trajectory adaptation \citep{mimicgen, softmimicgen} and visual generative models \citep{vace} by operating entirely in 3D space. By anchoring the target object with an explicit 3D mesh and propagating its 6D pose trajectory via the robot's forward kinematics, our approach ensures that the synthesized counterfactual demonstrations maintain both physical admissibility and multi-view consistency.

\begin{figure}[t]
    \centering
    \includegraphics[width=\linewidth]{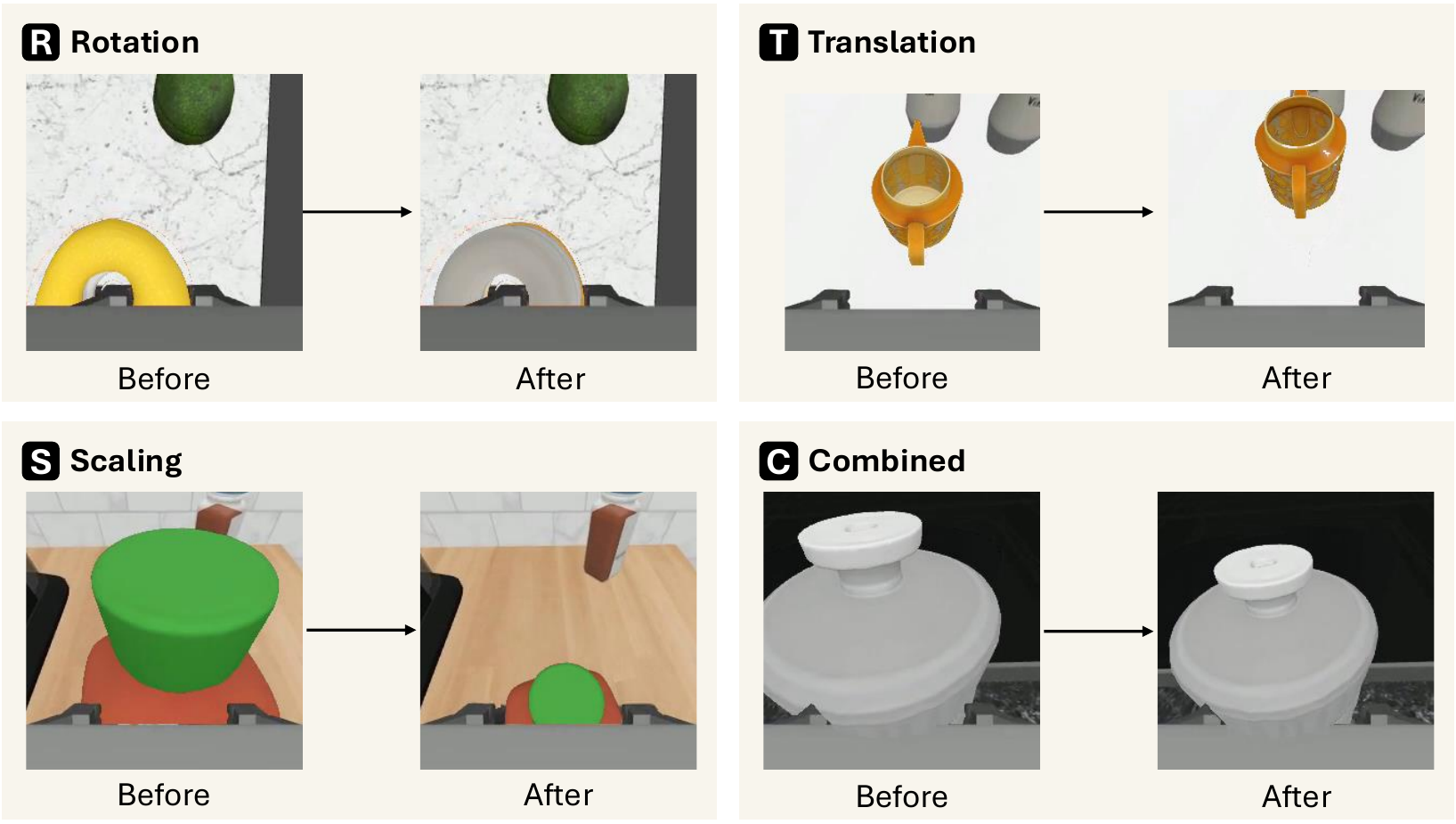}
    \caption{
    \textbf{Data augmentation details.} We apply combinations of geometric perturbations to the swapped target mesh, each shown as a \emph{before}~$\rightarrow$~\emph{after} pair.
    \textbf{(Top left)~Rotation:} the target mesh is randomly rotated or flipped to expose the policy to diverse plausible orientations.
    \textbf{(Top right)~Translation:} the mesh is shifted along the gripper's approach axis, varying the relative gripper--object offset.
    \textbf{(Bottom left)~Scaling:} the mesh is rescaled (\eg along the vertical axis) so that its size stays within a graspable range relative to the gripper.
    \textbf{(Bottom right)~Combined:} the above perturbations are composed.
    All perturbations remain grounded in 3D, keeping the augmented object in a physically plausible, graspable configuration after substitution.
    }
    \label{fig:detailed_method}
\end{figure}

\section{Method}
\label{sec:method}

Given a successful manipulation episode with a source object, our goal is to replace the source object with a target object across all camera views while preserving the physical plausibility of the recorded trajectory. Rather than applying view-independent generative editing, which struggles with multi-view consistency and out-of-distribution egocentric views, we operate directly in 3D.

We target three core constraints for physically consistent augmentation across multiple camera views: (1) the replacement object must follow a geometrically accurate 3D trajectory that reflects the original manipulation, (2) the replacement must be conditioned on the observed grasp so that the resulting interaction remains physically plausible, and (3) the edited object must appear consistently across all camera views and timesteps. To ensure multi-view consistency by construction, we represent the target object as an explicit 3D mesh and render it at a shared world-frame pose for each camera---rather than performing independent per-view edits that would require post-hoc alignment.

Our pipeline decomposes object-centric augmentation into three stages: target mesh reconstruction and 6D pose extraction (\cref{subsec:pose}), object augmentation (\cref{subsec:substitution}), and 3D mesh-pose guided video composition (\cref{subsec:composition}).

\paragraph{Input and output.} We take as input RGB videos from $N$ calibrated cameras $\{C_i\}_{i=1}^{N}$ with known intrinsics $K_i$ and extrinsics $[R_i \mid t_i]$, per-timestep object poses, the end-effector trajectory and gripper state (open/closed) over time, per-frame object masks, a source object mesh $\mathcal{M}_s$, and a target object mesh $\mathcal{M}_t$ reconstructed via \texttt{SAM3D}. In simulation, these inputs are read from the simulator state. On real robot data, camera parameters are obtained by stereo calibration with a ChArUco board, object masks are predicted by \texttt{SAM~2}~\citep{sam2}, and object poses are estimated as described in \cref{subsec:pose}, while the end-effector trajectory and gripper state are recorded during teleoperation. The output is a set of augmented multi-view videos in which the source object is replaced by the target object while preserving the robot's original motion, grasp contact, and cross-view geometric consistency.

\subsection{Target Mesh Reconstruction and 6D Pose Extraction}
\label{subsec:pose}

\paragraph{Target mesh reconstruction.} We reconstruct 3D meshes of candidate target objects using an off-the-shelf image-to-3D model~\citep{sam3d}. Importantly, we do not use the simulator's ground-truth, metrically scaled asset meshes (\eg the native RoboCasa365~\citep{robocasa365} assets) directly; we instead assume the realistic setting in which only an observation of the target object to be augmented is available. To emulate this, we render the target object to a single image and pass it to \texttt{SAM3D}, which returns the reconstructed mesh $\mathcal{M}_t$ used by our pipeline; the ground-truth mesh enters only as a stand-in for this observation and is never used directly. We further assume that both the source mesh $\mathcal{M}_s$ and the reconstructed target mesh $\mathcal{M}_t$ reside in the same aligned canonical coordinate frame, following~\cite{mimicgen}. This shared canonical space allows us to substitute one object for another without requiring per-instance pose re-estimation: any pose applied to the source mesh can be directly transferred to the target mesh after the augmentations described in \cref{subsec:substitution}. Because the target object thereby inherits the source object's pose trajectory in this shared frame, the gripper--object contact configuration of the original grasp is preserved by construction, up to the rescaling introduced in \cref{subsec:substitution}.

\paragraph{6D pose extraction.} Rather than treating each video frame as an independent editing target, we represent the manipulated object's motion as a continuous sequence of 6D poses in a shared world coordinate frame. This formulation ensures that edits are spatially grounded and temporally coherent across views.

Let $T^w_o(t) \in SE(3)$ denote the object pose in the world frame at timestep $t$. In simulator-based benchmarks and test benches \citep{robocasa365, libero}, ground-truth object poses are readily available from the environment state. On real robot data, we estimate the pose at the first frame and propagate it with the recorded robot kinematics, which removes the need for per-frame pose tracking under gripper occlusion. We initialize $T^w_o(0)$ with an RGB-D pose estimator~\citep{any6d} on stereo depth and refine it with point-to-point ICP and mask-IoU maximization. Before the grasp, the object rests on the supporting surface, so its pose stays fixed. After the gripper closes at $t_g$, the object is rigidly coupled to the parallel-jaw gripper and follows the relative end-effector motion:
\begin{equation}
    T^w_o(t) = T^w_e(t)\,\big(T^w_e(t_g)\big)^{-1}\,T^w_o(t_g), \quad t \geq t_g,
    \label{eq:propagation}
\end{equation}
where $T^w_e(t)$ is the recorded end-effector pose, and the object's vertical axis is constrained to remain perpendicular to the supporting surface. Our pipeline is agnostic to the source of the pose signal; it requires only a sequence of $SE(3)$ poses aligned to the world frame.

To obtain the object pose in each camera's coordinate frame, we apply the known camera extrinsics:
\begin{equation}
    T^{c_i}_o(t) = T^{c_i}_w \, T^w_o(t),
    \label{eq:camera_projection}
\end{equation}
where $T^{c_i}_w$ is the world-to-camera transform for camera $C_i$. Since all cameras reference the same world-frame trajectory, multi-view consistency of the object's spatial position and orientation is maintained by construction.

\subsection{Object Augmentation}
\label{subsec:substitution}

To increase the diversity of training data and improve generalization, we apply three augmentation strategies to the object meshes during data generation (\cref{fig:detailed_method}):
\begin{itemize}[leftmargin=*]
\item \textbf{Rotation.} We randomly rotate or flip the object mesh to expose the model to a variety of plausible orientations and poses, encouraging robustness to different object configurations during grasping.
\item \textbf{Translation.} We translate the object mesh closer to or farther from the gripper along the approach axis, enabling the model to learn from a wider range of relative gripper--object distances and grasp offsets.
\item \textbf{Scaling.} Although objects within the same category share similar geometry, their dimensions (\eg height and width) can vary considerably. For instance, after swapping one bottle mesh for another, the replacement may be too tall or too short relative to the gripper to yield a feasible grasp. To address this, we scale the object mesh by a factor $s$ along the vertical axis such that its height remains within a graspable range relative to the gripper's position, ensuring physically plausible contact after substitution. We denote the resulting rescaled mesh by $s \cdot \mathcal{M}_t$.
\end{itemize}

\subsection{3D Mesh-Pose Guided Video Composition}
\label{subsec:composition}

Given the recovered pose trajectory and the scaled target mesh, we compose the augmented video through geometry-guided rendering rather than per-frame generative editing. This design ensures that multi-view and temporal consistency are maintained by construction: every camera observes the same 3D mesh rendered at the same world-frame pose, eliminating the need for cross-view or cross-frame regularization.

\paragraph{Background completion.} For each frame and camera view, we first remove the source object from the original video using the provided object masks. The masked regions are then filled with a video inpainting model~\citep{void} to produce a clean background plate. Importantly, generative methods are used only for this background region completion. The appearance and position of the replacement object are determined entirely by geometric rendering.

\paragraph{Object rendering and compositing.} At each timestep $t$ and for each camera $C_i$, we render the scaled target mesh using the object pose $T^{c_i}_o(t)$ from \cref{eq:camera_projection} and the camera intrinsics $K_i$. The rendered object is then composited onto the inpainted background following a strict depth-ordered layering: the clean background plate is placed first, the target object rendering is overlaid next, and finally the robot and gripper regions from the original frame are restored on top using their segmentation masks. This back-to-front compositing preserves correct occlusion ordering, ensuring, for example, that gripper fingers remain visible in front of the grasped object, and produces frames in which the robot--object spatial relationship is physically consistent across all views.

\paragraph{Multi-view consistency.} Because all camera views render the same 3D mesh $s \cdot \mathcal{M}_t$ at a shared world-frame pose $T^w_o(t)$, the appearance of the target object is geometrically consistent across views by construction. This stands in contrast to generative video editing approaches, which would require explicit multi-view constraints or struggle to maintain identity and geometry of the replacement object across cameras.

\section{Experiments}
\label{sec:exp}

To evaluate the effectiveness of our augmentation pipeline, we test whether \method can improve a state-of-the-art robot foundation model on realistic household manipulation tasks. Our experiments are designed to answer three questions: (1)~Does augmentation recover failures when combined with the original data? (2)~Can augmented data alone teach manipulation of hard objects, both in simulation and on a real robot? (3)~How much does each augmentation component contribute?

\begin{table}[t]
\centering
\small
\caption{\textbf{Overall performance on failure episodes.}
Success rate (\%) and turnover ratio (\%) evaluated on failure episodes: instances where the base policy never succeeded in completing the task with the target object. Results are split into in-distribution (ID) object categories and out-of-distribution (OOD) object categories unseen during training, and averaged over 3 independent seeds. Higher is better for both metrics. \textbf{Bold} indicates the best result.
}
\label{tab:main}

\begin{NiceTabular}{l cc ccc ccc}
    \toprule
    & \multicolumn{2}{c}{Properties} & \multicolumn{3}{c}{Success Rate (\%)} & \multicolumn{3}{c}{Turnover Ratio (\%)} \\
    \cmidrule(lr){2-3} \cmidrule(lr){4-6} \cmidrule(lr){7-9}
    \multirow{2}{*}{Method} & Multi-view & \multirow{2}{*}{Sim-free} & \multirow{2}{*}{ID} & \multirow{2}{*}{OOD} & \multirow{2}{*}{Avg.} & \multirow{2}{*}{ID} & \multirow{2}{*}{OOD} & \multirow{2}{*}{Avg.} \\
    & Consistency & & & & & & & \\
    \midrule
    Base Policy & - & - & \zero & \zero & \zero & \zero & \zero & \zero \\
    MimicGen \citep{mimicgen}
      & \cmark & \xmark & 14.7 & 17.3 & 15.8 & 16.4 & 18.1 & 17.2 \\
    VACE \citep{vace}
      & \xmark & \cmark & 12.8 & 21.2 & 16.4 & 14.9 & 22.4 & 18.2 \\
    \midrule
    \rowcolor{green!10}
    \textbf{\method (Ours)}
      & \cmark & \cmark & \textbf{21.2} & \textbf{24.7} & \textbf{22.8} & \textbf{24.6} & \textbf{24.2} & \textbf{24.5} \\
    \bottomrule
\end{NiceTabular}

\end{table}

\subsection{Task Setup and Baselines}
\label{subsec:setup}

\paragraph{Benchmark and task.}
We evaluate \method both in simulation, on RoboCasa365~\cite{robocasa365}, and on a real robot. RoboCasa365 is a large-scale simulation benchmark for household mobile manipulation that features diverse everyday tasks on a wide range of environmental setups and objects, including assets from Objaverse \citep{objaverse}. RoboCasa365 provides the environmental diversity and object variation necessary to stress-test augmentation methods: policies must generalize across varied scene layouts as well as a wide range of object geometries within each object category. We focus on the \emph{Counter-to-Cabinet} task, a pick-and-place task in which the robot grasps an object from a kitchen counter and places it inside a cabinet. On a real robot, we evaluate \method and the baselines on a tabletop pick-and-place task, in which the robot moves an object on a table into a basket (\cref{fig:real_task}). These tasks are representative of common household manipulation and involve multi-step coordination (approach, grasp, transport, placement), making them suitable testbeds for evaluating whether augmented data can improve manipulation of diverse object instances.

\paragraph{Baselines.}
We compare \method against two augmentation baselines that represent fundamentally different strategies for generating additional training data:

\begin{itemize}[leftmargin=*,itemsep=2pt]
    \item \textbf{VACE}~\cite{vace}: A unified video creation and editing framework based on a diffusion transformer. Given a source video and an inpainting mask, VACE generates an edited video in which the masked region is filled with the target object specified via a reference image and text prompt. We apply VACE independently to each camera view.
    \item \textbf{MimicGen}~\cite{mimicgen}: A trajectory-level data generation system that adapts a small set of source demonstrations to new scene configurations via $SE(3)$ subtask transformations executed in simulation. Only transformed trajectories that complete the task when executed in simulation with the target object are retained. This filtering ensures physical plausibility, but yields no data for object--scene pairs without a successful rollout, and requires full access to the simulator and the target object asset.
\end{itemize}

\paragraph{Real-world setup.}
\label{subsec:real}
Our real robot follows the DROID setup~\citep{droid}: a Franka Research 3 7-DoF arm with a Robotiq 2F-85 parallel-jaw gripper, observed by one exocentric and one wrist-mounted ZED stereo camera (\cref{fig:real_hw}). We collect 30 teleoperated demonstrations per object, which \method augments with randomized poses. We evaluate four target objects: a Bref spray bottle, a snack box, an aluminum can, and a keyring-sized deformable doll, each augmented from source episodes of an object in the same category whose shape and appearance differ substantially (\cref{fig:real_pairs}). Depth for pose initialization is obtained from the ZED neural stereo depth. We evaluate each object over 30 trials, placing it at varying positions across the tabletop. We compare against VACE, since MimicGen requires simulator rollouts and is not applicable to real data, and report the success rate and class recovery (\cref{subsec:metrics}).

\begin{table}
\centering
\small
\caption{\textbf{Effect on hard examples.}
    We report the overall success rate (\%) in the top block, obtained with the model fine-tuned on the 8 most challenging unseen, out-of-distribution object instances, evaluated on 20 pick-and-place episodes per instance (\textbf{Bold} indicates the best result),
    and the class recovery per hard example in the bottom block (\cmark~if successful recovery, \xmark~otherwise). Object names correspond to mesh identifiers in the RoboCasa365~\citep{robocasa365} asset library.
}
\label{tab:hard}
\begin{NiceTabular}{l c c c c}
    \toprule
     ~~~~Model & Base Policy & MimicGen \citep{mimicgen} & VACE \citep{vace} & \cellcolor{green!10}\textbf{\method (Ours)}  \\
 Success Rate & 15/160 (9.4\%) & 9/160 (5.7\%) & 24/160 (15.0\%) & \cellcolor{green!10}\textbf{34/160} (\textbf{21.2}\%) \\
\midrule
{Donut5} & - & \xmark & \cmark & \cellcolor{green!10}\cmark \\
{Jar023} & - & \cmark & \cmark & \cellcolor{green!10}\cmark \\   
{MeasuringCup009} & - & \xmark & \xmark & \cellcolor{green!10}\cmark \\  
{SoapDispenser010} & - & \xmark & \xmark & \cellcolor{green!10}\cmark \\    
{Steak8} & - & \xmark & \cmark & \cellcolor{green!10}\cmark \\ 
{SyrupBottle005} & - & \cmark & \xmark & \cellcolor{green!10}\xmark \\    
{Teapot6} & - & \xmark & \cmark & \cellcolor{green!10}\cmark \\ 
{Teapot7} & - & \xmark & \cmark & \cellcolor{green!10}\cmark \\ \midrule
\end{NiceTabular}
\end{table}

\paragraph{Implementation details.}
\label{subsec:implementation}

We use \texttt{GR00T-1.5}~\cite{groot} as the base robot foundation model for all experiments. For the simulation experiments, we train the model for 30K steps on the combined dataset of original Counter-to-Cabinet demonstrations and augmented episodes generated by each respective method (\cref{tab:main}). For the hard-example experiments (\cref{tab:hard}), we initialize from the checkpoint fine-tuned on the original demonstrations and perform an additional 1.5K fine-tuning steps using only augmented episodes targeting hard examples. For the real-robot experiments, we train the base policy for 15K steps on collected human demonstrations, followed by an additional 6K fine-tuning steps on the generated augmented episodes. All fine-tuning runs use AdamW~\cite{adamw} with a learning rate of $1 \times 10^{-5}$ and cosine learning rate decay. For the VACE baseline, we use \texttt{Wan2.1-VACE-1.3B}, which we found to produce more plausible edits than its other variants. Each simulation fine-tuning run uses a single NVIDIA A100 (80 GB) GPU and takes approximately 8 GPU-hours. Our augmentation pipeline is fully automated and requires no human intervention, taking 117.6\,s on average to generate one augmented episode in RoboCasa365 on a single A100.

\begin{figure}[t]
    \centering
    \includegraphics[width=\linewidth]{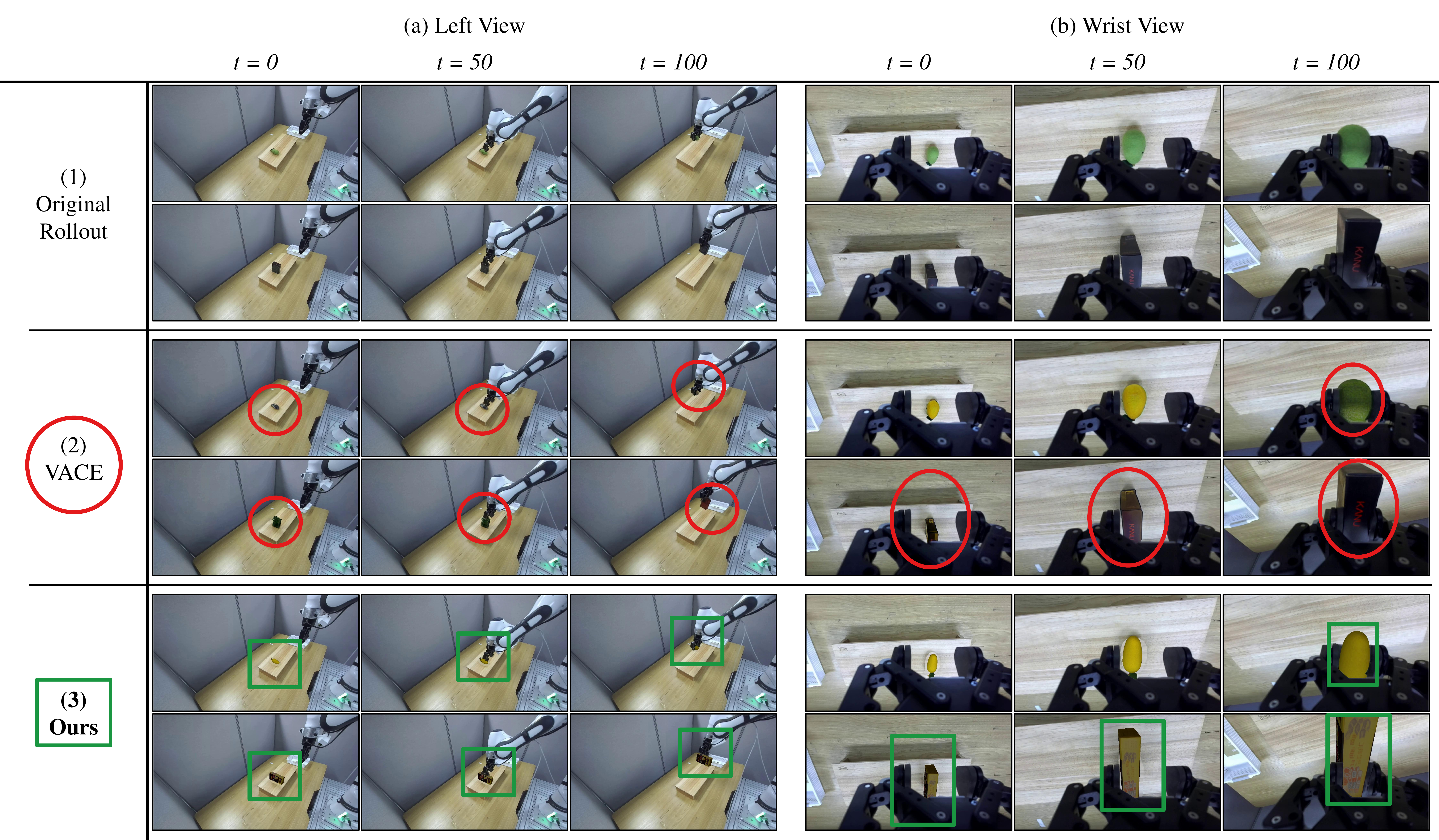}
    \caption{
    \textbf{Qualitative comparisons on real robot data.} We compare the original rollouts, the augmentations by VACE~\citep{vace}, and those by our \method in the left (exocentric) and wrist (egocentric) views at $t = 0, 50, 100$, for two source--target pairs (top: pear-shaped doll $\rightarrow$ mango-shaped doll; bottom: coffee box $\rightarrow$ snack box). {\color{red}Red ellipses} highlight inconsistencies of the object edited by VACE, which processes each view independently, such as the source object reappearing in the wrist view at $t = 100$. {\color{green!60!black}Green boxes} mark the same regions for \method, where the target object remains consistent across views and over time.
    }
    \label{fig:qualitative}
\end{figure}

\subsection{Augmentation Protocol}
\label{subsec:augmentation}

\paragraph{Identifying augmentation targets.}
To focus augmentation on the instances where it is most needed, we first characterize which object meshes are difficult for the base policy. We perform 200 rollouts per checkpoint. From the failure cases, we identify meshes that fail most frequently as well as meshes entirely unseen during training. For each candidate mesh, we then generate new episodes using a pick-and-place task across multiple scene configurations to verify that the difficulty is systematic rather than an artifact of a single layout. Through this procedure, we identify a set of 8 meshes---object instances on which the base policy achieves $\leq$30\% success rate---as the target objects for augmentation.

\paragraph{Pairing and augmentation.}
We pair each identified hard example with source episodes from the original Counter-to-Cabinet training set, constraining pairs to objects within the same semantic category (\eg replacing one cereal box with another). For each source--target pair, we generate 16 augmented episodes by randomly sampling rotation and translation perturbations, while scaling is applied when the target object's height falls below the gripper end-effector to ensure a feasible grasp. The same pairing and quantity are used for the VACE baseline to ensure a controlled comparison. For MimicGen, generating a physically valid trajectory is not always feasible for a given scene initialization; we therefore allow 32 attempts per pair and retain only successful rollouts.  This protocol yields 176 augmented episodes for both \method and VACE. MimicGen, due to rollout failures on certain object--scene combinations, produces only 44 valid episodes.

\subsection{Metrics}
\label{subsec:metrics}

We report three metrics, evaluated over 200 rollouts per configuration for the main evaluation and 20 rollouts per instance for the hard-example evaluation (\cref{tab:hard}):

\begin{itemize}[leftmargin=*,itemsep=2pt]
    \item \textbf{Success Rate:} The fraction of rollouts in which the robot successfully completes the task.
    \item \textbf{Turnover Ratio:} Among the set of meshes that never succeeded under the base checkpoint, the fraction that achieve at least one success after training with augmented data. This metric captures whether augmentation enables the policy to handle previously intractable instances, beyond aggregate success rate improvements.
    \item \textbf{Class Recovery:} A per-instance indicator of whether a specific object instance that consistently failed under the base policy achieves at least one successful rollout after fine-tuning. While the turnover ratio aggregates this signal into a single scalar, class recovery provides a fine-grained view of which object instances each method is able to recover.
\end{itemize}

\begin{table}[t!]
\centering
\small
\caption{\textbf{Real-world results.} We report the success rate (\%) on tabletop pick-and-place with 4 unseen target objects in the top block, each evaluated over 30 trials at varying positions on the tabletop (\textbf{Bold} indicates the best result), and the class recovery per object in the bottom block (\cmark~if recovered, \xmark~otherwise). Base Policy is trained only on the collected human demonstrations, and VACE and \method further fine-tune it on their augmented episodes.}
\label{tab:real}
\begin{NiceTabular}{l c c c}
    \toprule
    & Base Policy & VACE \citep{vace} & \cellcolor{green!10}\textbf{\method (Ours)} \\
    \midrule
    Success Rate & 10.0 & 8.3 & \cellcolor{green!10}\textbf{25.0} \\
    \midrule
    Bref Spray & - & \cmark & \cellcolor{green!10}\cmark \\
    Snack Box & - & \cmark & \cellcolor{green!10}\cmark \\
    Aluminum Can & - & \xmark & \cellcolor{green!10}\cmark \\
    Doll & - & \xmark & \cellcolor{green!10}\cmark \\
    \bottomrule
\end{NiceTabular}

\vspace{2mm}
\caption{\textbf{Ablation.} Contribution of each augmentation component, measured by the success rate (\%) on the 8 hard examples of \cref{tab:hard} with 20 episodes per instance. \emph{Object swap only} substitutes the target mesh without geometric augmentation, and \emph{Geometric aug.\ only} applies the rotation, translation, and scaling of \cref{subsec:substitution} to the source object without swapping. Each variant fine-tunes the base policy exclusively on its own augmented episodes. \textbf{Bold} indicates the best result.}
\label{tab:ablation}
\begin{NiceTabular}{l c}
    \toprule
    Method & Success Rate \\
    \midrule
    Base Policy & 9.4 \\
    Object swap only & 13.8 \\
    Geometric aug.\ only & 15.6 \\
    \midrule
    \rowcolor{green!10}
    \textbf{\method (Ours)} & \textbf{21.2} \\
    \bottomrule
\end{NiceTabular}
\end{table}

\subsection{Experimental Results}
\label{subsec:results}

We structure our evaluation around the three questions above, which together characterize the practical value of each augmentation method.

\paragraph{Q1: Does augmentation recover failures when combined with the original data?}
We first examine whether appending augmented episodes to the original training set can turn previously failed rollouts into successes. \cref{tab:main} reports success rates and turnover ratios evaluated exclusively on episodes where the base policy originally failed. \method achieves the highest average success rate of 22.8\%, outperforming both VACE (16.4\%) and MimicGen (15.8\%). The turnover ratio tells a similar story: \method converts 24.5\% of previously intractable instances into at least one successful rollout, compared to 18.2\% for VACE and 17.2\% for MimicGen. Notably, VACE-augmented data yields limited gains and in some configurations degrades performance, which we attribute to multi-view inconsistencies in the generated videos that introduce conflicting visual signals during training.

\paragraph{Q2: Can augmented data alone teach manipulation of hard objects, both in simulation and on a real robot?}
To isolate augmentation quality from the original training data, we fine-tune the policy exclusively on augmented episodes and evaluate on hard, unseen objects. In RoboCasa365, we evaluate on the 8 hardest out-of-distribution meshes selected following the protocol in \cref{subsec:augmentation} (\cref{tab:hard}). \method achieves a 21.2\% success rate, outperforming VACE (15.0\%) and MimicGen (5.7\%), and successfully recovers 7 out of 8 hard objects, compared to 5 and 2, respectively. On the real robot, we evaluate on the four target objects in \cref{subsec:real} (\cref{tab:real}). \method improves the base policy's success rate from 10.0\% to 25.0\% and recovers all four objects, including a deformable doll, using estimated poses, predicted masks, and reconstructed meshes. In contrast, VACE achieves 8.3\% and recovers only two objects. This indicates that \method's augmentations alone can improve manipulation of unseen objects, highlighting the importance of 3D-grounded, multi-view consistent augmentation. \cref{fig:qualitative} further illustrates that VACE produces inconsistent object appearances across views and timesteps, whereas \method maintains multi-view and temporal consistency. Additional simulation examples, including MimicGen, are provided in \cref{app:qualitative_sim}.

\paragraph{Q3: How much does each augmentation component contribute?}
To measure the contribution of each component, we repeat the hard-example experiment with object swap only and with geometric augmentation only, \ie the rotation, translation, and scaling of \cref{subsec:substitution} applied to the source object without swapping (\cref{tab:ablation}). Object swap alone raises the success rate of the base policy from 9.4\% to 13.8\%, geometric augmentation alone raises it to 15.6\%, and combining both reaches 21.2\%. The two components are complementary: object swap supplies the target object as a mesh-grounded, multi-view consistent substitution, and the geometric augmentation, enabled by the explicit mesh under a tracked $SE(3)$ trajectory, turns each swap into a diverse set of executable demonstrations.

\paragraph{Discussion.}
MimicGen relies on successful rollouts, which creates a coverage bottleneck: valid episodes exist only for a subset of mesh--scene combinations, and the resulting data can even degrade the base policy (\cref{app:full_results}). As shown in \cref{fig:qualitative_sim}, its adapted trajectories fail to produce valid grasps for challenging geometries, leaving no usable data for these instances. VACE avoids this bottleneck by editing in image space, but editing each view independently causes identity drift across frames, inconsistent shape and color across the left, right, and wrist cameras, and views in which the source object is left unedited (\cref{fig:qualitative,fig:qualitative_sim}); these discrepancies propagate as noisy supervision during training. In contrast, \method renders a single 3D mesh at a shared world-frame pose along a verified successful trajectory, so the replacement object keeps a consistent identity, geometry, and contact across egocentric and exocentric views by construction. It thus inherits the scalability of generative editing, as any source--target pair can be augmented, while preserving the physical grounding of simulation-based methods, and extends the policy to previously failed object instances without sacrificing in-distribution performance.

\section{Conclusion}
\label{sec:conclusion}

In this paper, we introduced {\method}, a 3D-aware data augmentation framework designed to enhance the generalization of robotic manipulation policies. To address the high cost of multi-view data collection, our approach leverages a pose-aware object swapping pipeline that integrates 3D mesh extraction, video inpainting, and robot kinematics. By operating in 3D space rather than purely in pixels, \method ensures that synthesized demonstrations maintain both physical admissibility and multi-view consistency. 
Our experimental evaluation on the RoboCasa365 benchmark and on a real robot demonstrates that \method improves the performance of VLA policies on challenging and novel object instances. Our method outperforms 2D video editing and simulation-based trajectory adaptation baselines, proving the effectiveness of our augmentation pipeline.
Furthermore, the inclusion of geometric perturbations, such as object rotations, translations, and rescaling, proved critical for improving out-of-distribution generalization.
We believe that grounding visual augmentation in 3D geometry is a promising direction for scaling robot learning across diverse real-world environments.

\paragraph{Limitations.}
\label{sec:limitations}
Several limitations remain. First, the object substitution process is naturally constrained by grasp compatibility; replacing a source object with a target object of drastically different scale or geometry can lead to physically implausible grasps. Automated affordance matching would be a valuable extension to handle broader categories. Second, \method{} heavily relies on the quality of external vision models, such as 3D mesh extractors and video inpainting networks. Consequently, it inherits the failure modes of these foundation models (\eg inpainting artifacts or incomplete meshes) and its visual fidelity is tied to their continued improvement. Finally, directly rendering 3D meshes into 2D inpainted backgrounds may not perfectly capture complex lighting interactions, such as dynamic cast shadows or advanced reflections. Modeling the interaction between the swapped object and the scene and incorporating neural relighting techniques are promising directions for future work.

\newpage
\begin{ack}
This work was partly supported by Institute of Information \& Communications Technology
Planning \& Evaluation (IITP) grant funded by the Korea government (MSIT)
(No.~RS-2019-II190075, Artificial Intelligence Graduate School Support Program (KAIST);
No.~RS-2022-II220959, Few-shot Learning of Causal Inference in Vision and Language for
Decision Making; IITP-2026-RS-2023-00255968, Artificial Intelligence Convergence Innovation Human Resources
Development) and National IT Industry Promotion Agency (NIPA) grant funded by the
Korea government (MSIT) (AI-Based Dark Factory Design-Deployment Solution Development, No.~RS-2026-25621604).
This research was also supported by the ``Advanced GPU Utilization Support Program'' funded by the
Government of the Republic of Korea (Ministry of Science and ICT).
We are grateful to RLWRLD Inc.\ for generously providing compute resources and
supporting the real-world robot experiments conducted in this work.
\end{ack}

{\small
\bibliography{ref}
\bibliographystyle{unsrtnat}
}

\clearpage
\appendix
\crefalias{section}{appendix}
\section{Appendix}
\label{sec:appendix}

\subsection{Full Evaluation on All Episodes}
\label{app:full_results}

\begin{table}[h]
\centering

\label{tab:appendix_full}
\begin{NiceTabular}{l ccc}
    \toprule
    & \multicolumn{3}{c}{Success Rate (\%)} \\
    \cmidrule(lr){2-4}
    Method & ID Object & OOD Object & Avg. \\
    \midrule
    Base Policy & 52.2 & 36.9 & 47.2 \\
    MimicGen \citep{mimicgen}
      & 52.2 & 33.3 & 46.0 \\
    VACE \citep{vace}
      & 51.2 & 40.4 & 47.7 \\
    \midrule
    \rowcolor{green!10}
    \textbf{Pose6DAug (Ours)}
      & \textbf{52.7} & \textbf{42.9} & \textbf{49.5} \\
    \bottomrule
\end{NiceTabular}
\caption{Overall success rate (\%) evaluated on all episodes, including both previously successful and previously failed instances. Bold indicates the best result.}

\end{table}

\cref{tab:main} in the main text reports performance exclusively on failure episodes. To assess whether augmentation preserves the base policy's existing capabilities, we evaluate each method on the \emph{full} set of 200 rollouts per configuration, which includes both previously successful and previously failed instances. \cref{tab:appendix_full} reports the overall success rate across all episodes.

Notably, MimicGen's overall performance (46.0\%) falls below that of the base policy (47.2\%), indicating that its augmented data degrades the model's ability to handle objects it could already manipulate. VACE maintains near-baseline performance but provides only modest gains on OOD objects. In contrast, \method achieves the highest overall success rate (49.5\%) while improving OOD performance from 36.9\% to 42.9\%, demonstrating that our augmentation strengthens generalization without sacrificing in-distribution competence.

\subsection{Qualitative Comparisons in Simulation}
\label{app:qualitative_sim}
\begin{figure}[t]
    \centering
    \includegraphics[width=\linewidth]{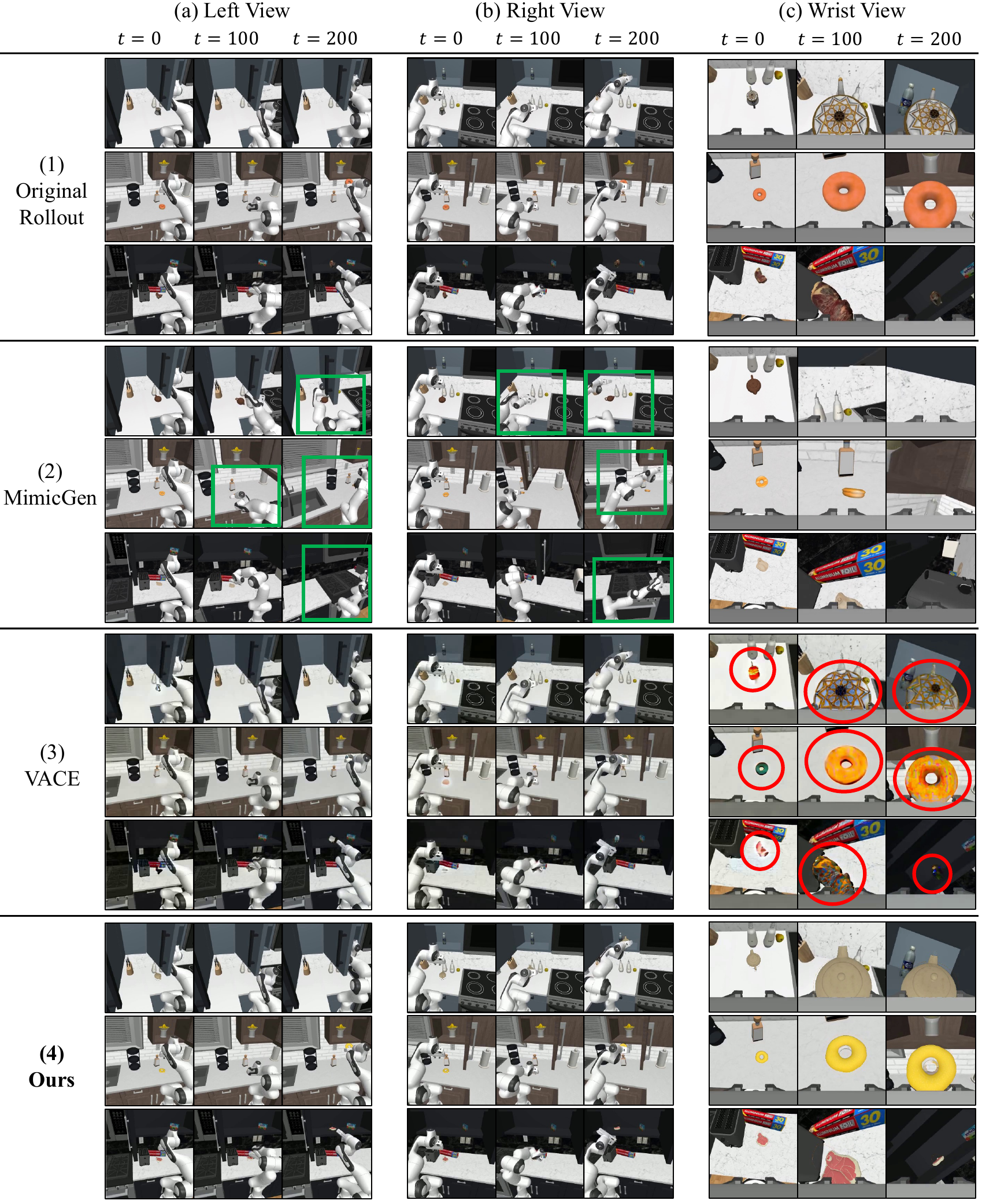}
    \caption{
    \textbf{Qualitative comparisons in simulation.} We compare the baselines \citep{mimicgen, vace} and our \method across left (exocentric), right (exocentric), and wrist (egocentric) camera views. {\color{red}Red circles} highlight multi-view inconsistencies of the augmented object, and {\color{green!70!black}green rectangles} indicate the physical implausibility of the action trajectory.
    }
    \label{fig:qualitative_sim}
\end{figure}

\cref{fig:qualitative_sim} compares MimicGen, VACE, and \method on RoboCasa365 across the left, right, and wrist camera views at $t=0$, $100$, and $200$, together with the original rollout from which each augmentation is generated. Each row corresponds to a different source episode and target object. For MimicGen, on the challenging set of meshes, the proposed trajectories fail to grasp the object (green rectangles), so only a small fraction of rollouts succeed and MimicGen produces few physically plausible demonstrations. VACE preserves the original robot motion but edits each camera view independently, and therefore lacks multi-view consistency. As highlighted by the red circles, the appearance of the generated object changes over time in the wrist view and does not match its appearance in the exocentric views, where the source object is in some cases left unedited. In contrast, \method keeps the original trajectory and renders a single target mesh at a shared world-frame pose in every view, so the target object keeps the same identity and geometry across all views and throughout the episode, and stays in the gripper during transport.

\subsection{Real-World Experimental Setup}
\label{app:real_setup}
\begin{figure}[h]
    \centering
    \includegraphics[width=0.7\linewidth]{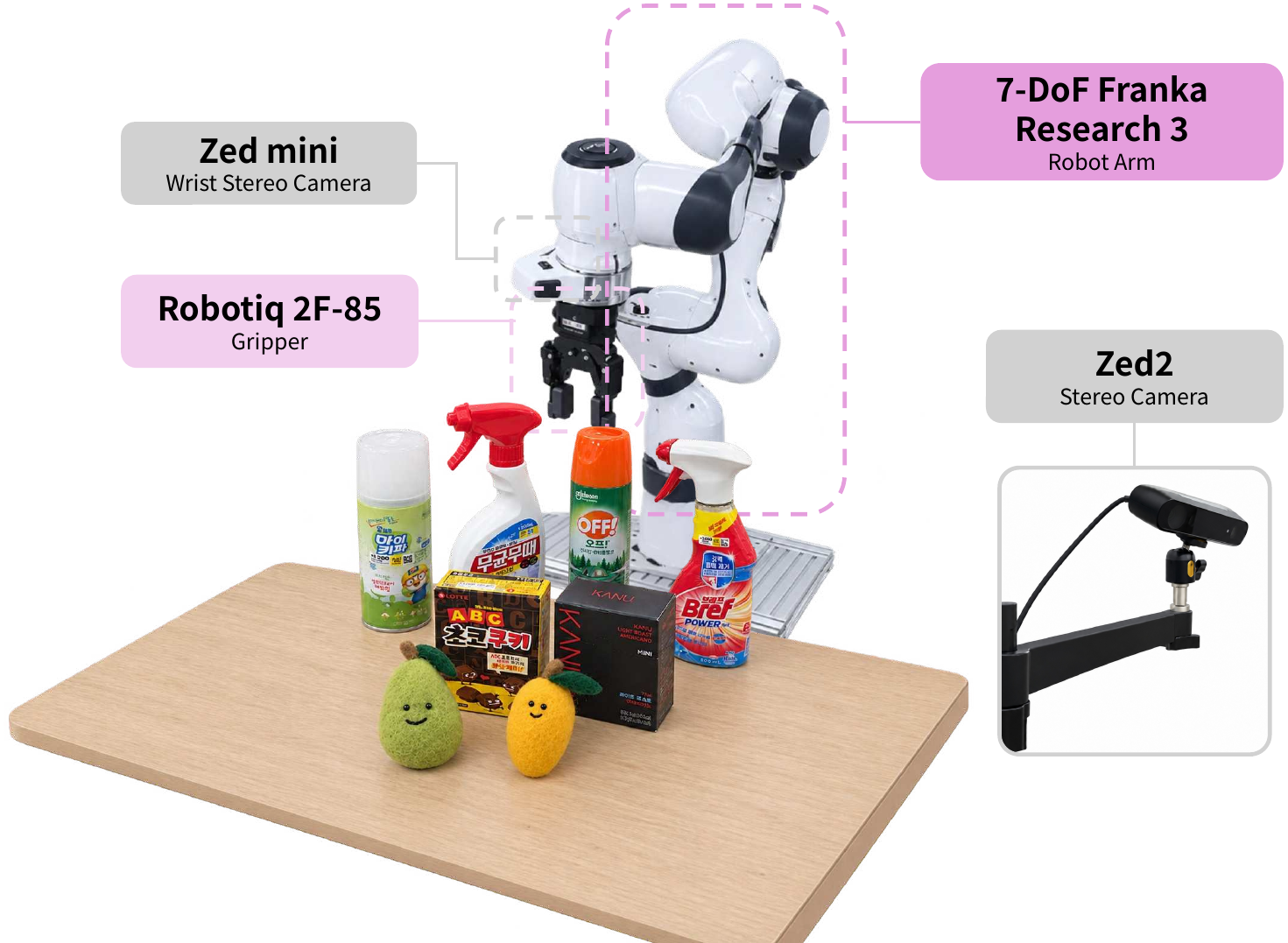}
    \caption{\textbf{Real-robot hardware setup.} The objects used in our real-world experiments are shown on the table.}
    \label{fig:real_hw}
\end{figure}

\begin{figure}[h]
    \centering
    \includegraphics[width=\linewidth]{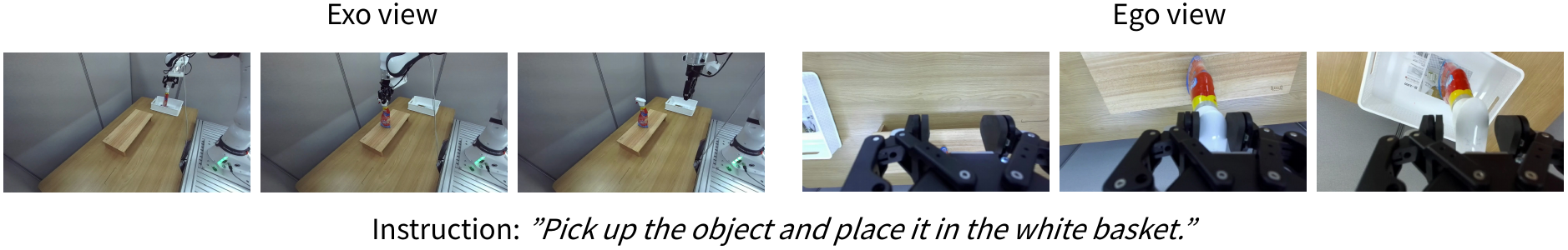}
    \caption{\textbf{Real-world task.} Example frames of a pick-and-place demonstration from the exocentric (left) and wrist-mounted egocentric (right) cameras.}
    \label{fig:real_task}
\end{figure}

\begin{figure}[h]
    \centering
    \includegraphics[width=\linewidth]{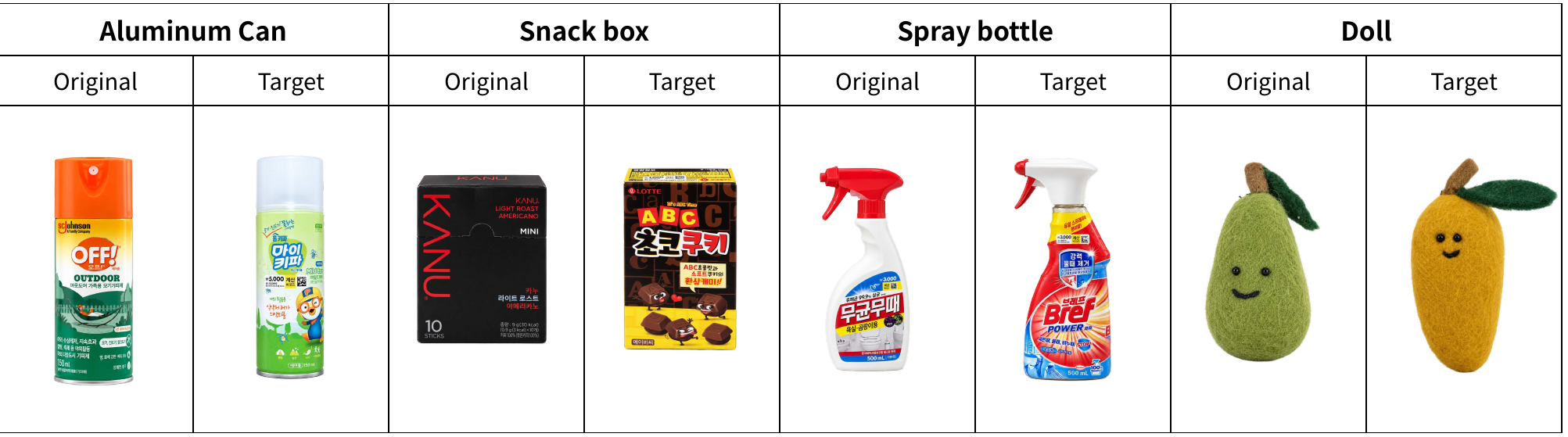}
    \caption{\textbf{Source--target pairs for real-world augmentation.} For each category, the source object (Original) in the collected demonstrations is replaced with a target object (Target) whose shape and appearance differ substantially.}
    \label{fig:real_pairs}
\end{figure}

\cref{fig:real_hw,fig:real_task,fig:real_pairs} show the hardware, example frames of the task, and the source--target pairs of the real-world experiments in \cref{subsec:real}.

\paragraph{Hardware.} The robot is a 7-DoF Franka Research 3 arm with a Robotiq 2F-85 parallel-jaw gripper (\cref{fig:real_hw}). A ZED~2 stereo camera on a fixed mount provides the exocentric view, and a ZED~Mini stereo camera on the wrist provides the egocentric view. Both cameras are calibrated with a ChArUco board, which provides the intrinsics and extrinsics used to render the target mesh into each view.

\paragraph{Task and demonstrations.} Given the instruction ``Pick up the object and place it in the white basket,'' the robot grasps the object on the table and places it in the basket (\cref{fig:real_task}). We collect 30 demonstrations per object by teleoperation, recording both camera streams together with the end-effector trajectory and gripper state, and \method augments them with randomized poses. The base policy is trained on these demonstrations, and VACE and \method each fine-tune it on their augmented episodes (\cref{subsec:implementation}).

\paragraph{Augmentation pipeline.} Object masks are predicted by \texttt{SAM~2}~\citep{sam2}. The source object pose is initialized at the first frame with Any6D~\citep{any6d} on the ZED neural stereo depth, refined with ICP, and propagated with the recorded end-effector poses after the grasp (\cref{eq:propagation}). Each target object mesh is reconstructed with \texttt{SAM3D}~\citep{sam3d}.

\paragraph{Source--target pairs.} \cref{fig:real_pairs} shows the four categories. In each pair, the source object appears in the collected demonstrations, while the target object is unseen during base policy training. The target differs from the source in shape, size, color, and texture: an aluminum can with a different body and cap, a snack box with a different aspect ratio and print, a spray bottle with a different trigger head and bottle shape, and a felt doll with a different body shape and color.

\paragraph{Evaluation protocol.} Each target object is evaluated over 30 trials, placing it at varying positions across the tabletop, for a total of 120 trials per method. We report the success rate over all trials and, for each object, class recovery (\cref{subsec:metrics}), \ie whether the fine-tuned policy succeeds at least once.

\subsection{Failure Cases}
\label{app:failurecase}
\begin{figure}[h]
    \centering
\includegraphics[width=\linewidth]{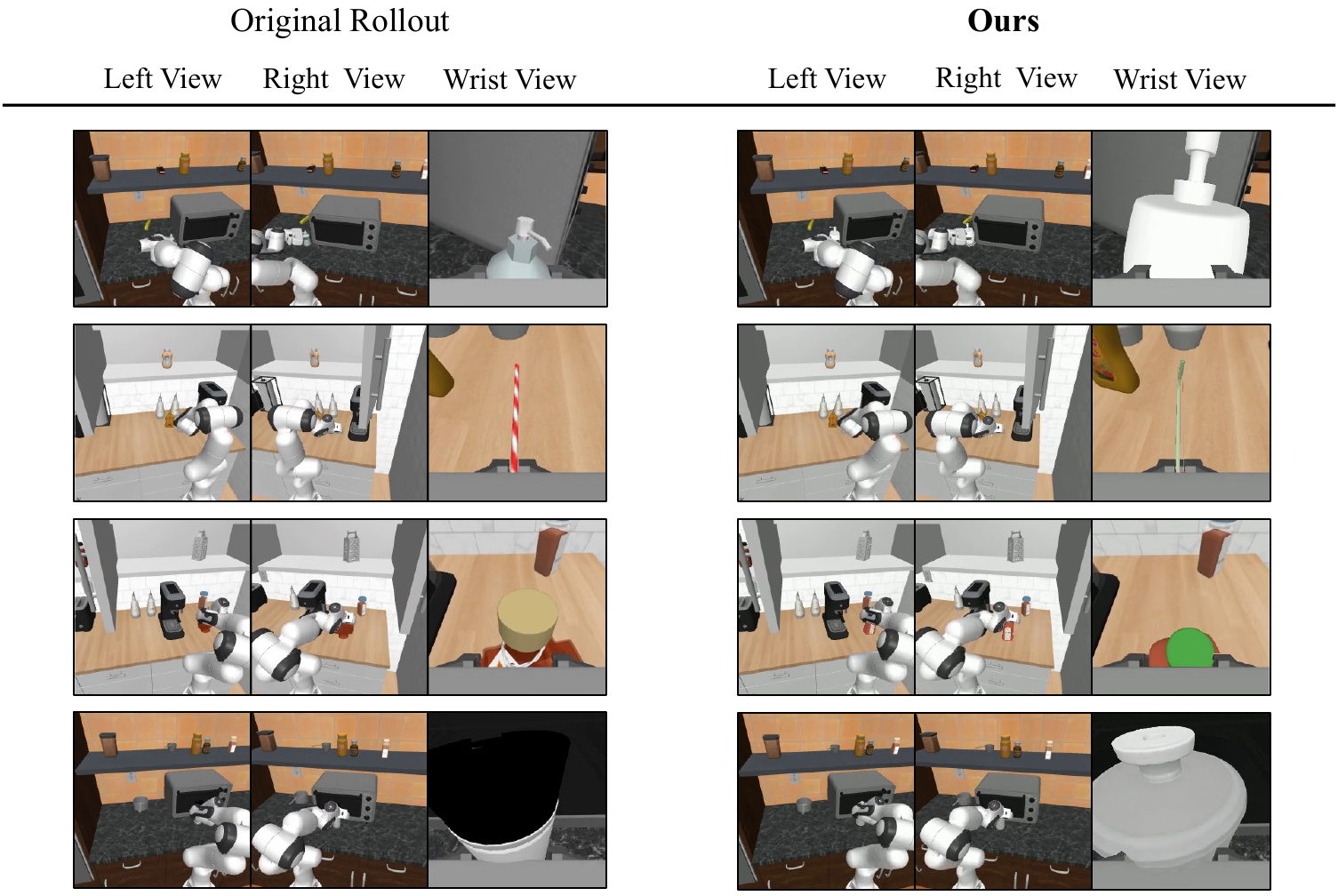}
    \caption{
    \textbf{Failure cases of \method.} 
    }
    \label{fig:failure}
\end{figure}

We show failure cases of our pipeline in \cref{fig:failure}. These failures are primarily driven by discrepancies in scale, width, or centroid position between the source and target objects. For instance, when the object is substantially wider, the rendered mesh may intersect with the gripper or extend beyond the graspable region.

\paragraph{Failure boundary.} These failures arise in the object augmentation stage (\cref{subsec:substitution}). Since the source trajectory fixes the gripper--object contact configuration, a swap remains physically plausible only while the grasp required by the target object stays within the demonstrated grasp distribution. Failure occurs when the contact configuration itself must change, \eg when the grasp axis must be reassigned, which no rigid perturbation of the target mesh can recover. To quantify this boundary, we perturb each augmentation axis beyond the range used in our protocol and test the physical plausibility of the result. Rotation preserves the contact geometry and remains plausible over the largest deviations we tested ($\pm30^\circ$ tilt and $\pm45$--$135^\circ$ yaw). Translation and scaling directly alter the gripper--object contact: translation beyond the range seen in the demonstrations ($\pm10$\,mm) breaks plausibility in 24 of 56 cases (42.9\%), and scaling away from the grasp-fitted value ($\times0.8$--$0.9$) breaks it in all 8 cases. Accordingly, \method samples rotations broadly while keeping translation and scale within the grasp-compatible range.

\subsection{Potential Societal Impact}
\label{app:societal-impact}
While our data augmentation design can be beneficial for various robotic applications, such as object manipulation and tool use, the emergence of unexpected behavior within~\method~can lead to misrepresentations of the safe action data. 
For those applications that require extremely accurate models for safety-related judgments, such as policy learning for assistive humanoid robots for people with disabilities or children, the unexpected behaviors must be carefully managed. 
To ensure the reliability of systems using our method,
we recommend conducting thorough investigations and implement robust mitigation strategies to minimize potential risks, thereby increasing the overall safety and effectiveness of these applications.

\end{document}